\documentclass[conference]{IEEEtran}
\IEEEoverridecommandlockouts
\usepackage{cite}
\usepackage{amsmath,amssymb,amsfonts}
\usepackage{graphicx}
\usepackage{textcomp}
\usepackage{xcolor}
\usepackage{algorithm}
\usepackage{algorithmicx}
\usepackage{algpseudocode}
\usepackage{color}
\usepackage{booktabs}
\usepackage{verbatim}
\def\BibTeX{{\rm B\kern-.05em{\sc i\kern-.025em b}\kern-.08em
    T\kern-.1667em\lower.7ex\hbox{E}\kern-.125emX}}
\begin{document}

\title{Adapting Rules of Official International Mahjong for Online Players}

\author{\IEEEauthorblockN{ Chucai Wang }
\IEEEauthorblockA{\textit{School of Computer Science} \\
\textit{Peking University}\\
Beijing, China \\
wangchucai@pku.edu.cn}

\and
\IEEEauthorblockN{ Lingfeng Li}
\IEEEauthorblockA{\textit{School of Computer Science} \\
\textit{Peking University}\\
Beijing, China \\
Lingfengli@stu.pku.edu.cn}

\and
\IEEEauthorblockN{ Yunlong Lu}
\IEEEauthorblockA{\textit{School of Computer Science} \\
\textit{Peking University}\\
Beijing, China \\
luyunlong@pku.edu.cn}

\and
\IEEEauthorblockN{ Wenxin Li}
\IEEEauthorblockA{\textit{School of Computer Science} \\
\textit{Peking University}\\
Beijing, China \\
lwx@pku.edu.cn}
}

\maketitle

\begin{abstract}

As one of the worldwide spread traditional game, Official International Mahjong can be played and promoted online through remote devices instead of requiring face-to-face interaction. However, online players have fragmented playtime and unfixed combination of opponents in contrary to offline players who have fixed opponents for multiple rounds of play. Therefore, the rules designed for offline players need to be modified to ensure the fairness of online single-round play. Specifically, We employ a world champion AI to engage in self-play competitions and conduct statistical data analysis. Our study reveals the first-mover advantage and issues in the subgoal scoring settings. Based on our findings, we propose rule adaptations to make the game more suitable for the online environment, such as introducing compensatory points for the first-mover advantage and refining the scores of subgoals for different tile patterns. Compared with the traditional method of rotating positions over multiple rounds to balance first-mover advantage, our compensatory points mechanism in each round is more convenient for online players. Furthermore, we implement the revised Mahjong game online, which is open for online players. This work is an initial attempt to use data from AI systems to evaluate Official Internatinoal Mahjong's game balance and develop a revised version of the traditional game better adapted for online players.


\end{abstract}

\begin{IEEEkeywords}
Mahjong,  game design, champion AI
\end{IEEEkeywords}

\section{Introduction}

While traditional games can be played on the internet with the help of information technologies, some of them require rule adaptations for better game balance due to the properties of online environments. For example, the limited game rounds may enhance the unfairness of first-mover advantage. In this work, we select Official International Mahjong, a standard variant of mahjong which is a four-player traditional game with millions of players worldwide. The rules of Official International Mahjong were designed for offline players. Although there is asymmetry in different positions during a single match, the score is treated as the same among all positions. Also, the subgoal scoring system was created based on traditional human experience and lack of further discussions on the frequency and mathematical combinations of tile patterns. Our work further discuss the balance of this traditional game in one-round case and proposes some rule adaptions to provide a better experience for online players.

To evaluate the game balance, earlier methods usually analyze game data collected by many test players and get expert opinions on game rules, which costs both in time and human expertise~\cite{b1}. However, the development of game AI has opened up new opportunities to assess the balance of traditional games. For example, self-play matches of AlphaGo ~\cite{b2} are significantly biased toward White, indicating that the fair value of komi should be less than the value currently adopted in mainstream rules of Go, which had been summarized from human players' data when the rule was designed. 

Recently, the holding of the Mahjong AI Competition in IJCAI~\cite{b3} makes it possible to assess the game balance of this traditional game with AI programs. This competition attracts players from universities, companies, and the community of Mahjong. AI programs submitted by these participants with different algorithms, and the winner shows state-of-the-art performance under the rules of Official International Mahjong. Match data is generated from this champion AI in the online arena of Botzone~\cite{b4}, making it possible to analyze this game's game balance.

Our work analyzes the self-play games of the champion AI in this competition. We propose rule adaptions from two aspects: a compensation mechanism for first-mover advantage and adaptating  the scores of subgoals for different tile patterns. These adaptions turn Official International Mahjong into a game that only require one round to play which is suitable for online players. Furthermore, We developed this revised Mahjong game on the internet which is open for online players. Our work is an initial attempt to improve the game balance of this popular game, Official International Mahjong for one-round case and develop a revised version to be better adapted for online players.

\section{Related Work}

\subsection{Online Game}

With the development of IT technology, the medium of games has expanded from physical to the virtual electronic world. Digital games serve people in entertainment, education~\cite{b5} and  research~\cite{b6}. People can access the digital game through remote devices and face-to-face interactions were not necessary for players.

Online players have many different characteristics from offline players. Firstly, their opponents are unfixed and vary in each game. For instance, players can switch into another desk in an online poker game and their opponents can leave the game table when a game ends. Secondly, online players have fragmented playtime rather than offline players who have full time to complete a match. Based on these, the rules of online games will differ from those of offline games to ensure fairness.

\subsection{AI for Game Balance}

There are related work also utilizing AIs to improve game balance. Mark et al.~\cite{b7} firstly introduces AIs in GGP engine in the process of game design. Silva et al.~\cite{b8} claims that AI can serve as an evaluator for games. Christoffer et al.~\cite{b9} adopt MCTS algorithm as procedural personas for the testing of games. Deepmind researchers~\cite{b10} proposes AlphaZero as a framework to explore the balance of alternative rule sets in Chess. Joe et al.~\cite{b11} also uses restricted self-play of AIs to improve game rules. 

However, these work aim to create a more balanced game version and they modify the games directly. Our work is different from these because we aim to proposes rule adaptions to make this game more suitable for the online environment. And in techniques, these work focus games with perfect information, while Mahjong is a game with imperfect information, making it harder to train strong AIs through the AI framework of these work. Our work uses strong AIs from game AI competitions as black boxes to assess game balance.

\subsection{Analysis of Mahjong Rules}

There are related works conducting mathematical analysis on Mahjong rules. Cheng et al. \cite{b12} first studied a combinatorial problem in this game. They defined a 13-tile combination as a k-gate ($1 \leq k \leq 9$) combination if there are k tiles of different values such that each tile can transfer these 13 tiles into a winning hand. Furthermore, they compute how many different tile combinations for k-gate tiles. Li and Yan \cite{b13} propose a notion of deficiency to evaluate the quality of 14 tiles and present a policy to discard tiles.

Our analysis of Mahjong is different from their work in several ways. We statistic the data from a top AI in competitions, which gave us direct information on the performance of high-level players. And our analysis aims to improve the game's balance for online players rather than explore mathematical properties of the rules.

\subsection{Mahjong AI}

Mahjong proposes challenges to the game AI society for its hidden information. Zheng and Li \cite{b14} categorized Mahjong AI work into two research models: opponent modeling and self-perspective research. Currently, Suphx\cite{b15}, based on deep reinforcement learning, is the strongest AI for Japanese Mahjong, and it can beat most top human players on the Tenhou Website. In 2022, JueJong \cite{b16} solved 1-on-1 Official International Mahjong with a combination of CFR and actor-critic framework, beating a human champion. However, the four-player Official Internation Mahjong still needs to be solved.

Our work is different from theirs in the objectives. We use the Mahjong AI to analyze the balance of this traditional game rather than developing strong AI programs to defeat top human players. And we directly use state-of-art AIs as research tools.

\section{Background}

\subsection{Official International Mahjong }

\begin{figure}[tb]
  \centering
  \includegraphics[width=0.75\linewidth]{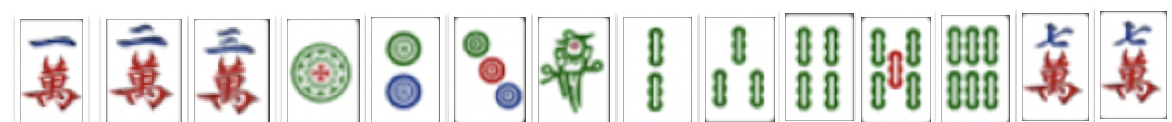}
  \caption{A common scoring pattern called "Mixed Triple Chow", including three same consecutive number sequences in all three suits.}
  \label{fig:mix_triple_chow}
\end{figure}

\begin{table}[tb]
  \caption{The complete table of scoring patterns in Official International Mahjong. There are 81 different patterns scoring from 1 points to 88 points.}
  \label{tab:tile_table}
  \begin{center}
  \begin{tabular}{|c|c|}\toprule
  \hline
    \textit{Pattern scores} & \textit{patterns}  \\ \midrule
    \hline
    1 point & Pure Double Chow, Mixed Double Chow, \\ & Short Straight, Two Terminal Chows, 
    \\ &Pung of Terminals or Honours, Melded Kong, \\ &One Voided Suit, No Honours, Edge Wait
    \\ &Closed Wait, Single Wait, Self-Draw \\
    \hline
    2 points & Dragon Pung, Prevalent Wind, Seat Wind, \\ &Concealed Hand, All Chows, Tile Hog,  \\ & Mixed Double Pung, Two Concealed Pungs, \\ &Concealed Kong, All Simples \\
    \hline
    4 points & Outside Hand, Fully Concealed Hand, \\ &Two Melded Kongs, Last Tile  \\
    \hline
    5 points & Melded and Concealed Kongs \\
    \hline
    6 points & All Pungs, Half Flush, Mixed Shifted Chows,  \\ &All Types, Melded Hand, Two Dragon Pungs  \\&Two Concealed Kongs\\
    \hline
    8 points & Mixed Straight, Reversible Tiles, \\ & Mixed Triple Chow, Mixed Shifted Pungs, \\ &Chicken Hand 
, Last Tile Draw, \\ & Out with Replacement Tile,  Rob Kong, \\ & Last Tile Claim  \\
\hline
    12 points & Lesser Honours and Knitted Tiles, \\ &Knitted Straight, Upper Four
, Lower Four, \\ &Big Three Winds \\
\hline
    16 points & Pure Straight, Three-Suited Terminal Chows, \\ & Pure Shifted Chows,  All Fives, Triple Pung,\\ &Three Concealed Pungs  \\
    \hline
    24 points & Seven Pairs,Lower Tiles   \\ &All Even Pungs, Full Flush, Pure Triple Chow,\\ & Pure Shifted Pungs, Upper Tiles, Middle Tiles, \\ &Greater Honours and Knitted Tiles\\
    \hline
    32 points & Four Pure Shifted Chows, Three Kongs, \\ &All Terminals and Honours  \\
    \hline
    48 points & Quadruple Chow, Four Pure Shifted Pungs \\
    \hline
    64 points & All Terminals, Little Four Winds, All Honours,\\ & Little Three Dragons, Pure Terminal Chows \\ &Four Concealed Pungs  \\ 
    \hline
    88 points & Big Four Winds, Big Three Dragons,  \\ &Nine Gates, Four Kongs, Seven Shifted Pairs, \\ & All Green, Thirteen Orphans  \\
    \hline
    \bottomrule
  \end{tabular}
  \end{center}
\end{table}

Official International Mahjong (Chinese Official Mahjong) is a rule set of Mahjong variants instituted by the General Administration of Sport in China in 1998. Since 2017~\cite{b17}, Official International Mahjong is admitted as part of the World Mind Sports Games and had professional ranking tournaments and international competitions. 

There are 144 tiles in a complete set of mahjong, 4 tiles for each kind. In a game, each player starts with 13 private tiles, draws tiles from tile walls, and chooses one to discard in turn. One can also grab other players' discarding tile instead of drawing a new tile when it can make a set with his private tiles. Such actions are called Pong, Chow, and Kong, depending on the type of tile sets. There are several tile patterns for scores as subgoals, and the first player to match patterns summing up to at least 8 points wins the game. Each player's score is calculated based on the points of the pattern and the source of the tile to win. If the winner wins $n$ points by drawing a tile, he gets a score of $n*3+24$ while all other players get $-n-8$. If he wins $n$ points by grabbing another player's discarded tile, that player gets a score of $-n-8$, and he gets $n+24$ while the other two both get a $-8$.

Official International Mahjong has 81 different scoring subgoals, each defining a tile pattern and its score. Table~\ref{tab:tile_table} shows the complete table of scoring patterns, from the simplest 1-point patterns to the rarest 88-point patterns. And Fig.~\ref{fig:mix_triple_chow} shows a typical scoring pattern called Mixed Triple Chow, which includes three identical consecutive number sequences in all three suits and scores 6 points. In the gameplay, players have to decide which combination of tile patterns they can make and discard tiles according to the plan until he makes Mahjong, i.e., his hand matches patterns summing up to at least 8 points after he draws a tile or grabs others' discarding tiles.

\subsection{IJCAI Mahjong AI competition }

\begin{figure}[tb]
  \centering
  \includegraphics[width=0.75\linewidth]{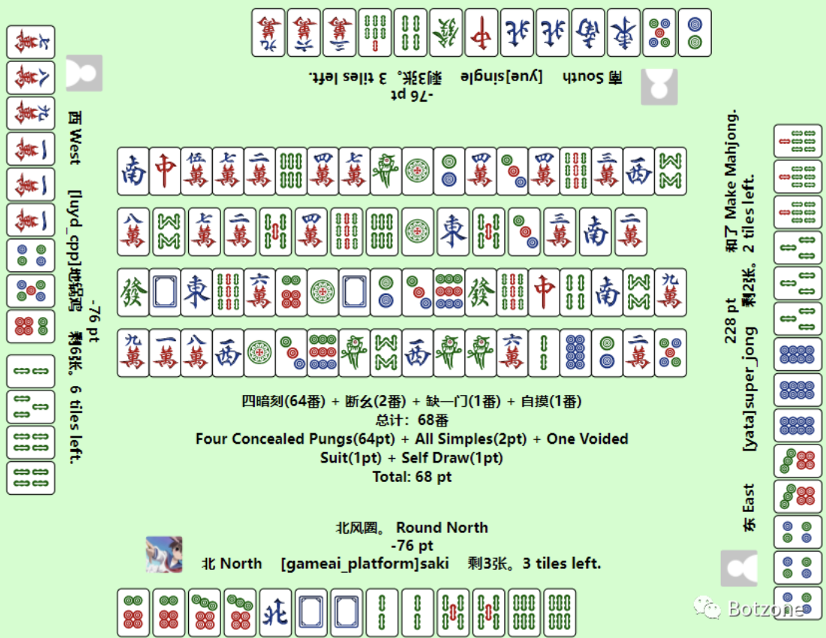}
  \caption{The Mahjong environment on Botzone.}
  \label{fig:botzone_mahjong}
\end{figure}

IJCAI 2022 Mahjong AI competition is held as a worldwide competition in IJCAI 2022 conference under Mahjong International League's guidance to promote AI development in this game. Many teams from universities, companies, and the community of Mahjong participated in this competition and built many high-level Mahjong AI programs. These programs are evaluated on Botzone, an online multi-agent competitive platform shown in Fig. ~\ref{fig:botzone_mahjong}, using a tournament system based on the Swiss System and a Duplicate Format of Mahjong. 

The competition is divided into three phases, the elimination round to decide the top 16, the semi-final round to select the top 4, and the final round to determine the champion. The champion AI turns out to be a high-level Mahjong agent that can beat average human players in the online arena of Botzone, and in the workshop, they claim that deep learning is used to train their AI agent.

\section{Rule Adaptments}

We evaluate the game balance of Official International Mahjong in two aspects, the first-mover advantage and the scores of tile patterns as subgoals, through the self-play games of the champion AI in the IJCAI 2022 Mahjong AI competition. The authors have approved the use of their AI programs and the generated game data. Based on the analysis of the game data, we further propose two methods to adaptate the rules of Mahjong for online players, the compensation mechanism to eliminate first-mover advantage and adapting the scores of  tile  patterns as subgoals.

\subsection{First-Mover Advantage}

\begin{figure}[tb]
  \centering
  \includegraphics[width=1.0\linewidth,height=0.66\linewidth]{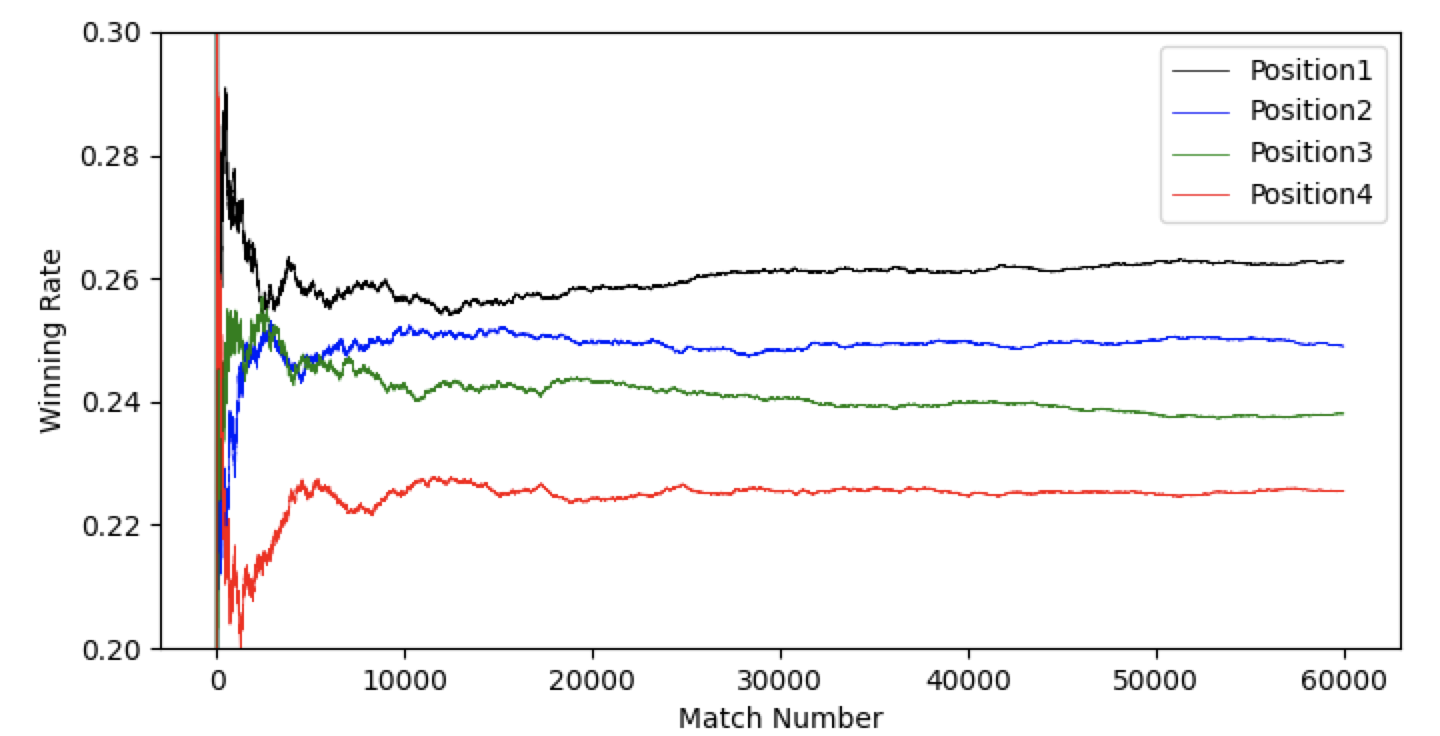}
  \caption{Average winning rates in the four positions}
  \label{fig:wr}
\end{figure}

\begin{figure}[tb]
  \centering
  \includegraphics[width=1.0\linewidth,height=0.66\linewidth]{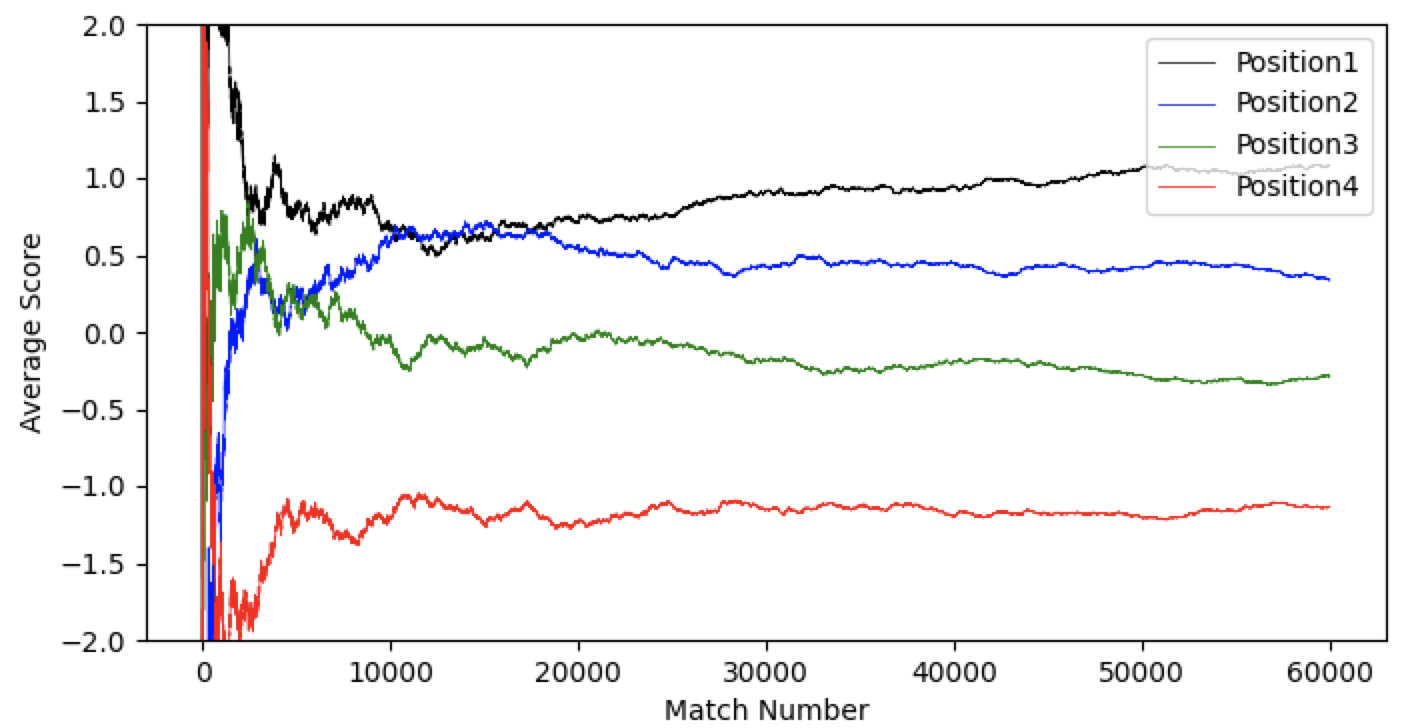}
  \caption{Average scores in the four positions}
  \label{fig:sc}
\end{figure}

\begin{table}[tb]
  \caption{The winning rate and average scores of the four positions in self-play games of the champion AI. The interval of 95\% confidence of the winning rate is also shown.}
  \label{tab:winrate}
  \begin{center}
  \begin{tabular}{|c|c|c|}\toprule
  \hline
    \textit{Position} & \textit{Winning Rates} & \textit{Average Scores} \\ \midrule
    \hline
    Position 1 & 26.19$\%$±0.11$\%$  & 1.0 \\
    \hline
    Position 2 & 25.21$\%$±0.11$\%$  & 0.4\\
    \hline
    Position 3 & 23.85$\%$±0.11$\%$  & -0.3\\
    \hline
    Position 4 & 22.45$\%$±0.11$\%$ & -1.1 \\ \bottomrule
    \hline
  \end{tabular}
  \end{center}
\end{table}

In asynchronous games where players take turns to perform actions, players may take advantage of the playing order to establish certain advantages, which break the game's balance. Theoretically, Zermelo \cite{b18} proved that for turn-based two-player games with perfect information and finite horizons, there are only three possible conditions: a winning strategy exists for the first or last player, or the optimal policies lead to a draw. But due to limited searching depth, game designers introduced komi values for the game of Go for a better game balance in practice. Similarly, there are rules of exchanging hands and forbidden hands for Gomoku.

As an asynchronous game, the first-mover advantage may also exist in Official International Mahjong. Still, it has not been thoroughly investigated because human players usually neglect or eliminate it by playing multiple rounds and adjusting the position of different players, making each player sit in each position at equal times. But the effect of first-mover advantage is still worth investigating in a single match for online players.

We use the champion AI to generate about 557,056 self-play games, randomly generating the initial hand and tile walls in each game. Fig.~\ref{fig:wr} and Fig.~\ref{fig:sc} show that agents' winning rates and average scores in the first 60,000 competitions differ among positions. The result of 557,056 self-play games are summarized in Table~\ref{tab:winrate} and shows the first player has about $3.74\%$ higher winning rate than the last player, which indicates a substantial first-mover advantage compared to other games, e.g., the two players in Go have a 0.5$\%$ \cite{b19} of difference in their winning rate.

To eliminate such first-mover advantage for online one-round play, we propose a compensation mechanism for Official International Mahjong inspired by the rule 'Komi' in the game of Go. Komi is a rule in Go that the first, i.e., the black player, needs to pay compensating points to the second, i.e., the white player, which is 3.75 in Chinese rule and 3.25 in Korean and Japanese rule. This rule of Komi makes the winning rate of the two players similar in high-level competitions by professional human players. Regarding Official International Mahjong, we also use a compensating mechanism where the first two players should give scores to the last two players in each match. We choose the compensating points based on the self-play data from the champion AI. Specifically, the first and second players pay 1.0 and 0.4 points, respectively, while the third and fourth players receive 0.3 and 1.1 points. Under such a compensation mechanism, it is expected that Mahjong players no longer need to switch their positions and play 4 games in each round for fairness of positions but can play an arbitrary number of games with fixed positions while keeping the fairness of the competition.

\begin{algorithm}[tb] 
	\renewcommand{\algorithmicrequire}{\textbf{Input:}}
	\renewcommand{\algorithmicensure}{\textbf{Output:}}
	\caption{Rank four AI agents without compensating points}
	\label{alg::std_exp} 
	\begin{algorithmic}[1] 
		\Require 
		Four AI agents.
		\Ensure 
		The average score of each agent per match.
		\State Initialize the accumulated scores of each agent as 0;
		\Repeat 
		\State Randomly generate fixed tile walls for each position;
		\For{$4!=24$ permutations of four agents}
		\State Run a match with specific player seating;
		\State Update the accumulated scores of each player by match result;
		\EndFor  
		\Until{A predefined number of rounds}
		\State Calculate the average score of each agent per match.
	\end{algorithmic} 
\end{algorithm}

\begin{algorithm}[tb] 
	\renewcommand{\algorithmicrequire}{\textbf{Input:}}
	\renewcommand{\algorithmicensure}{\textbf{Output:}}
	\caption{Rank four AI agents with compensating points}
	\label{alg::valid_exp} 
	\begin{algorithmic}[1] 
		\Require 
		Four AI agents.
		\Ensure 
		The average score of each agent per match before and after compensating points.
        \State Initialize the accumulated scores of each agent as 0.
		\State Fix the seating positions of four agents as one of the $4!=24$ possible permutations.
		\Repeat 
		\State Randomly generate tile walls for each position.
        \State Run a match with fixed player seating;
        \State Update the accumulated scores of each player by match result; 
		\Until{A predefined number of rounds}
        \State Calculate the average score of each agent per match;
		\State Adding compensating points to the average score of each player according to their seating positions.
	\end{algorithmic} 
\end{algorithm}

To validate that this compensation mechanism improves the balance of the original Mahjong rule, we adopt the top four AI agents from IJCAI 2022 Mahjong AI Competition and evaluate them under different competition formats. The authors have approved the use of these AI programs of them. Experiment 1 follows the same format as the original Mahjong AI Competition, which is the direct use of a Duplicate Format where four players switch their positions for a whole permutation of 24 matches in each round with the same initial hands and tile walls. A total of 1024 rounds are played in Experiment 1 to ensure convergence of the rankings, which is used as the 'correct' ranking of these four AI programs. In Experiment 2, four players take fixed positions and only play one match under each randomly generated initial hand and tile wall. The pseudocode of the competition format in both experiments is shown in Algorithm~\ref{alg::std_exp} and Algorithm~\ref{alg::valid_exp}.

\begin{table}[b]
  \caption{Average scores and rankings of the top 4 agents in Experiment 1 and Experiment 2 without compensation mechanism.}
  \label{tab:valid1}
  \begin{center}
  \begin{tabular}{|c|c|c|}\toprule
  \hline
    \textit{Agent} &\textit{Experiment 1}  & \textit{Experiment 2 (w/o compensation)}    \\ \midrule
    \hline
    Agent 1 & 0.835 (Rank 1)   & 3.323 (Rank 1)  \\
    \hline
     Agent 2 & 0.215 (Rank 2)  &-0.048 (Rank 2)  \\
     \hline
     Agent 3 & -0.422 (Rank 3)  &-1.705 (Rank 4)  \\
     \hline
    Agent 4  & -0.627 (Rank 4)  &-1.570 (Rank 3) \\ \bottomrule
    \hline
  \end{tabular}
  \end{center}
\end{table}

\begin{table}[b]
  \caption{Average scores and rankings of the top 4 agents in Experiment 1 and Experiment 2 with compensation mechanism.}
  \label{tab:valid2}
  \begin{center}
  \begin{tabular}{|c|c|c|}\toprule
  \hline
    \textit{Agent} &\textit{Experiment 1}  & \textit{Experiment 2 (with compensation)}  \\ \midrule
    \hline
    Agent 1& 0.835 (Rank 1) & 2.323 (Rank 1)\\
    \hline
     Agent 2& 0.215 (Rank 2)   & -0.448 (Rank 2)\\
     \hline
     Agent 3& -0.422 (Rank 3)   & -0.605 (Rank 3)\\
     \hline
    Agent 4& -0.627 (Rank 4)  & -1.270 (Rank 4) \\ \bottomrule
    \hline
  \end{tabular}
  \end{center}
\end{table}

Table~\ref{tab:valid1} shows the rankings and average scores of those AI programs in Experiment 2 compared with Experiment 1. The result shows that the rankings are inconsistent when using Duplicate Format to switch positions and using fixed positions, indicating that using fixed positions to evaluate agents can produce incorrect rankings with the interference of first-mover advantage. Table~\ref{tab:valid2} shows the rankings and average scores in Experiment 2 after we apply the compensation mechanism, producing a correct ranking compared with Experiment 1. This experiment shows that our compensation mechanism is necessary to evaluate agents' performance when their positions are not switched in a competition. It is worth noting that the fixed position used in Experiment 2 has 24 choices. We randomly take 4 of them to repeat Experiment 2, almost all producing the same rankings when applying the compensation mechanism, except for one case of inconsistent rankings of two players with very close scores. This further validates the robustness of our compensating mechanism. Detailed data is shown in Section I in Supplementary Material.

\subsection{Scores of tile patterns as subgoals}

\begin{figure*}[tb]
  \centering
  \includegraphics[width=1.0\linewidth,height=0.72\linewidth]{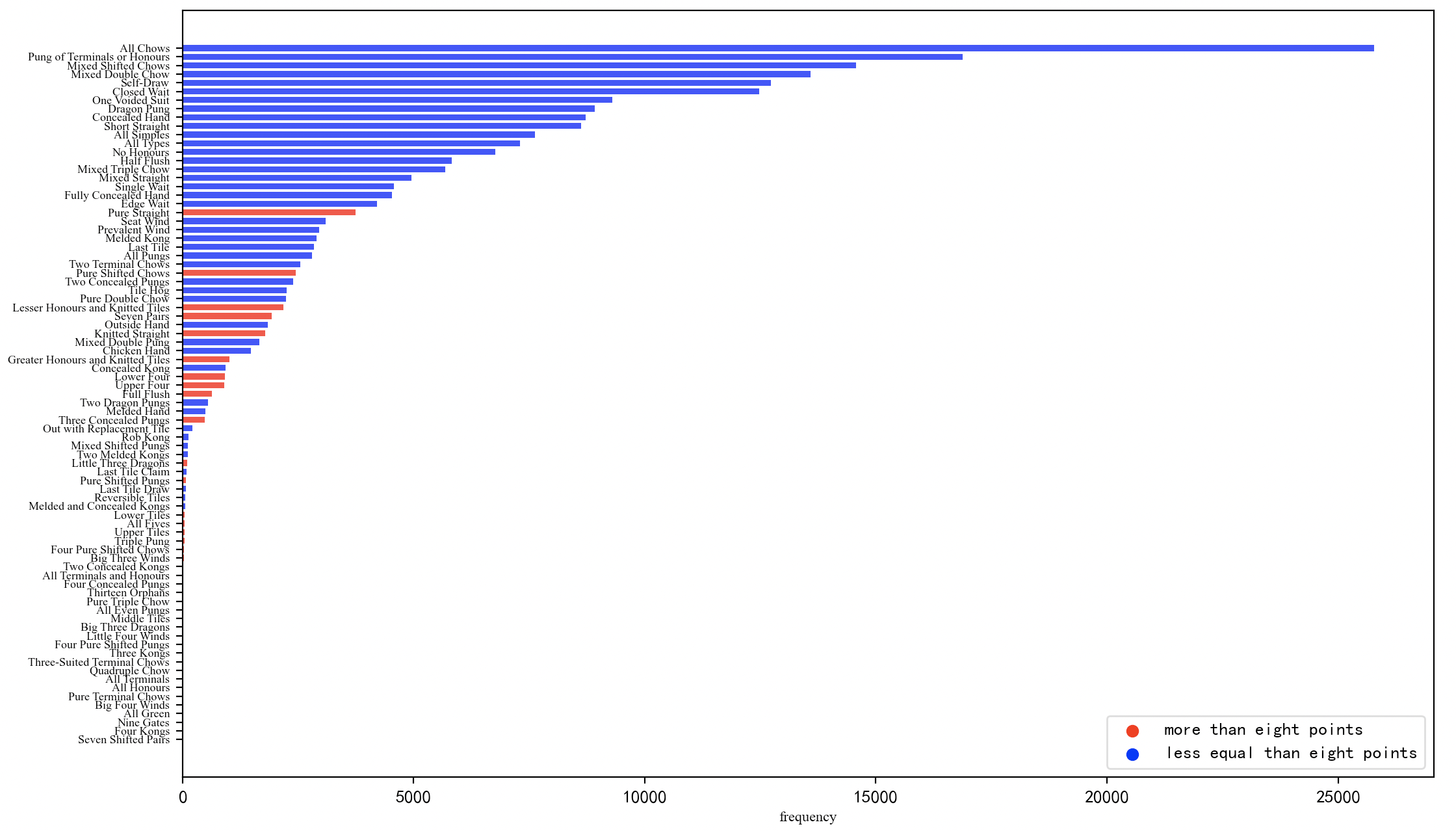}
  \caption{The number of occurrence of each scoring pattern in 65,536 self-play games from the champion AI.}
  \label{fig:frequen}
\end{figure*}

There are 81 different scoring patterns in the Official International Mahjong rule, as shown in Table~\ref{tab:tile_table}. The scores of these patterns served as subgoals for this game and were decided empirically when this rule set was initially designed. It is expected that rarer patterns should be assigned higher scores to encourage players to make them in one-round online play. However, people notice that some high-score patterns are common in gameplay, indicating that these patterns' scores were set high to influence the game's balance. For example, Seven Pairs is a frequent pattern when playing this game while it worth 24 points in Official International Mahjong ,which is considered high compared with other Mahjong variants such as Riichi Mahjong. Our work proposes a scoring adjustment draft based on the champion AI's self-play data.

We count the number of occurrences of each tile pattern in 557,056 self-play games of the champion AI. The result of the first 65,536 games indicates some common patterns are scoring high, as shown in Fig.~\ref{fig:frequen}. We notice that there are 43 tile patterns worth no more than 8 points in the original rule set, which should be those patterns that are easier to make. However, Table~\ref{tab:upper_41} shows the most frequently occurred 43 scoring patterns in the self-play data not only include patterns worth no more than 8 points but those patterns up to 24 points, indicating inconsistency between the frequency of each scoring patterns and their scores.

\begin{table}[tb]
  \caption{The most frequent 43 patterns in self-play games from the champion AI and their scores in the original rule set of Official International Mahjong.}
  \label{tab:upper_41}
  \begin{center}
  \begin{tabular}{|c|c|}\toprule
  \hline
    \textit{Scores} & \textit{Patterns}  \\ \midrule
    \hline
    1 point & Pure Double Chow, Mixed Double Chow, \\ & Short Straight, Two Terminal Chows,  
    \\ &Pung of Terminals or Honours, Melded Kong, \\ &One Voided Suit, No Honours, Edge Wait, 
    \\ &Closed Wait, Single Wait, Self-Draw \\
    \hline
    2 points & Dragon Pung, Prevalent Wind, Seat Wind, \\ &Concealed Hand, All Chows, Tile Hog,  \\ & Mixed Double Pung, Two Concealed Pungs, \\ &Concealed Kong, All Simples \\
    \hline
    4 points & Outside Hand, Fully Concealed Hand, Last Tile  \\
    \hline
    6 points & All Pungs, Half Flush, Mixed Shifted Chows,  \\ &All Types, Melded Hand, Two Dragon Pungs  \\
    \hline
    8 points &  Mixed Triple Chow, Chicken Hand,\\ &  Rob Kong, Mixed Straight    \\
    \hline
    12 points & Lesser Honours and Knitted Tiles, \\ &Knitted Straight, Upper Four, Lower Four \\
    \hline
    16 points & Pure Straight, Pure Shifted Chows \\
    \hline
    24 points & Seven Pairs, Greater Honours and Knitted Tiles, \\ & Full Flush \\ \bottomrule
    \hline
  \end{tabular}
  \end{center}
\end{table}

\begin{algorithm}[tb] 
	\renewcommand{\algorithmicrequire}{\textbf{Input:}}
	\renewcommand{\algorithmicensure}{\textbf{Output:}}
	\caption{Point adaption for scoring patterns}
	\label{alg::adjust} 
	\begin{algorithmic}[1]
		\Require
        The frequency of each scoring pattern in self-play games of Champion AI: Freq[];
		\Require
		The points of each scoring pattern: Point[];
		\Ensure 
		The improved points of each scoring pattern NewPoint[].
		\State Let PointLevel[] be the sorted list of points of all the scoring patterns in ascending order with duplicate number removed;
		\State Count the number of patterns whose points $ \leq 8$: N=43;
		\State Sort the scoring patterns based on their frequency of occurrence;
		\State Newpoint[] = Point[];
        \For{Pattern in the most frequent N patterns}
            \If{Point[Pattern] \textgreater 8}
                \State level = PointLevel.Index(Point[Pattern]);
                \State NewPoint[Pattern] = PointLevel[level - 1];
            \EndIf
        \EndFor
        \For{Pattern not in the most frequent $N$ patterns}
            \If{Point[Pattern] = 8}
                \State level = PointLevel.Index(8);
                \State NewPoint[Pattern] = PointLevel[level + 1];
            \EndIf
        \EndFor
        \State Return NewPoint[] as the improved points of each scoring pattern;
	\end{algorithmic} 
\end{algorithm}

\begin{table}[tb]
  \caption{Adaption draft for the points of scoring patterns}
  \label{tab:adjustment}
  \begin{tabular}{|c|c|c|}\toprule
  \hline
    \textit{pattern} & \textit{previous points} & \textit{new points}  \\ \midrule
    \hline
    Reversible Tiles & 8  & 12\\
    \hline
    Mixed Shifted Pungs & 8  & 12\\
    \hline
    Lesser Honours and Knitted Tiles & 12  & 8\\
    \hline
    Knitted Straight & 12  & 8\\
    \hline
    Upper Four & 12  & 8\\
    \hline
    Lower Four & 12  & 8\\
    \hline
    Pure Straight & 16  & 12\\
    \hline
    Pure Shifted Chows & 16  & 12\\
    \hline
    Seven Pairs & 24  & 16\\
    \hline
    Greater Honours and Knitted Tiles & 24  & 16\\
    \hline
    Full Flush & 24 & 16 \\ \bottomrule
    \hline
  \end{tabular}
\end{table}

\begin{table}[tb]
  \caption{Comparison of patterns among different AIs}
  \label{tab:frequens}
  \begin{center}
  \begin{tabular}{|c|c|c|}\toprule
  \hline
    \textit{rankings} & \textit{overlap of top 43 patterns} & \textit{overlap of adjustments}  \\ \midrule
    \hline
    Rank2 & 43/43  & 11/11\\
    \hline
    Rank4 & 43/43  & 11/11\\
    \hline
    Rank10 & 43/43  & 11/11\\
    \hline
    Rank12 & 43/43 & 11/11 \\ \bottomrule
    \hline
  \end{tabular}
  \end{center}
\end{table}

\begin{table}[tb]
  \caption{theoretical analysis of patterns' combinatorial numbers .}
  \label{tab:combina}
  \begin{center}
  \begin{tabular}{|c|c|}\toprule
  \hline
    \textit{Scores} & \textit{Patterns with combinatorial numbers' levels and adjusments}  \\ \midrule
    \hline
    8 points & Mixed Straight (11), Mixed Triple Chow (11), \\ & Reversible Tiles (8, up),  Mixed Shifted Pungs (8, up) \\
    \hline
    12 points & Lesser Honours and Knitted Tiles (11, low), \\ &Knitted Straight (11, low), Upper Four (9, low),\\& Lower Four (9, low),  Big Three Winds (8) \\
    \hline
    16 points & Pure Straight (11, low), Pure Shifted Chows (10, low), \\ &   All Fives (9), Triple Pung (8), \\ & Three-Suited Terminal Chows (8),\\ 
    \hline
    24 points & Seven Pairs (12, low), \\ & Greater Honours and Knitted Tiles (10, low)    \\ &  Full Flush (9, low), Pure Shifted Pungs (8), \\ &  All Even Pungs (7),  Upper Tiles (7), Middle Tiles (7), \\ & Lower Tiles (7), Pure Triple Chow (5)\\
    \hline
    \bottomrule
  \end{tabular}
  \end{center}
\end{table}

Thus, we propose a draft of rule adaptions based on patterns' frequency. The original rules of Official International Mahjong only include 43 scoring patterns whose points are no more than the winning threshold of 8 points. So, the top 43 frequent patterns should lower their initial score if it exceeds 12. And the pattern should increase its score to 12 if its initial score is 8 and its frequency ranking is less than 43. As a complementary procedure, we fix the points of some patterns directly induced by luck like Last Tile Draw, which is determined as 8 points as a bonus for the winning threshold of the game. Algorithm~\ref{alg::adjust} shows the pseudocode of the adjustment, and Table~\ref{tab:adjustment} shows the result of our adjustment draft.

We also investigate the most frequently occurring tile patterns using other AI programs in the competition to avoid the bias introduced by champion AI's inner preference for scoring patterns. We randomly choose four AIs developed with different algorithms, which rank 2, 4, 10, and 12 in the competition. Using these AIs to conduct self-plays, we statistic the top 43 most frequent patterns and conduct the rule adjustments based on Algorithm~\ref{alg::adjust}, which is the same procedure as champion AI. These results in Table~\ref{tab:frequens} show consistency between the top 43 most frequent patterns and adjusted points with champion AI's results, further confirming that our adjustments avoid the champion AI's inner preference for scoring patterns. Detailed data is shown in Section II in Supplementary Material.

We also compare the frequency statistics with theoretical results. There are total $C^{14}_{144}+C^{15}_{144}+C^{16}_{144}+C^{17}_{144}+C^{18}_{144}$ combinations among which we compute the number of possible combinations for each tile patterns. The results are shown in Table~\ref{tab:frequens}, and the number beside the pattern's name is the magnitude of combination numbers. For example, the number of  Mixed Triple Chow is about $10^{11}$, and we add the number 11 beside it. We also annotate the adjustment direction that “low” means we lower the scores and “up” means we increase the scores in the adjustments shown in Table \ref{tab:combina}. Our adjustments tend to lower the scores of patterns with more combinational numbers and increase the scores of patterns with less combinational numbers, which further validates our rule adjustments. 

Furthermore, we implement the revised Mahjong game in the Botzone arena named MahjongRevised. Online players can access the game and enjoy human-human competitions on the platform. Also, Players can choose to develop AI agents for this game and open human-AI competitions. It is an initial attempt to develop a revised version of the traditional game better adapted for online players.

\section{Conclusions}

In this work, we analyze the self-play games of strong AIs in Mahjong AI competitions and propose rule adaptions to make them more suitable for online environments. By counting the win rates of different positions in self-play matches of champion AI, we show that the first player wins about 3.74$\%$ more than the last player, marking a first-mover advantage for single round match in this four-player game. We propose a compensation mechanism to nullify the first two players' advantage by compensating fixed points of 1.0 and 0.4 to the last two players who receive 0.3 and 1.1. We find that the top 4 players of the IJCAI competition with fixed seats can be correctly ranked only when we add compensating points to their points, further validating our compensating mechanism.

Also, we explore the scores of subgoals which were set emperically. We count the frequency of scoring patterns in the self-play games to show the mismatchings of the difficulties and their score. Furthermore, we propose rule adaptions by changing the scores of 11 patterns with initial points above 8 to make it more reasonable for one-round match. We also consider the frequency of three other AIs in the top 16 rankings and the combinational numbers of related patterns, and the results adopt our adjustments. Based on these, we developed the revised version of Official International Mahjong in Botzone, named as MahjongRevised which is accessible for all online players.

With the emergence of more AI competitions, better utilization of strong AIs in these competitions is demanded. We use them for game design and hope our effort can inspire more data applications from various AI competitions. This is an initial attempt to access the game balance of Official International Mahjong quantitatively, and we also hope it can bring a deeper understanding of this game for online players.

\section*{Acknowledgment}

We acknowledge the participants in IJCAI competitions who provide their AIs for our research.

\vspace{12pt}

\end{document}